%% file: main.tex
\documentclass[11pt,a4paper]{article}
\usepackage[hyperref]{naaclhlt2019}
\usepackage{times}
\usepackage{latexsym}

\usepackage{url}

\usepackage{xcolor}
\usepackage{tikz}
\usepackage{tikz-qtree}
\usepackage{latexsym}
\usepackage{graphicx}			
\usepackage{times}
\usepackage{latexsym}
\usepackage{mathpartir}
\usepackage{pifont}

\usepackage{proof}
\usepackage{xspace}
\usepackage{multirow}
\usepackage{forest}
\usepackage{subcaption}

\usepackage{dsfont}
\usepackage{amsmath}
\usepackage{amssymb}
\usepackage{amsthm}
\usepackage{algorithm}
\newcommand{\namecite}[1]{\newcite{#1}}

\input{defs}

    \makeatletter

\def\@fnsymbol#1{\ensuremath{\ifcase#1\or *\or
   \mathsection\or \mathparagraph\or \|\or **\or \dagger\dagger
   \or \ddagger\ddagger \else\@ctrerr\fi}}

\aclfinalcopy 

\title{Learning to Stop in Structured Prediction for Neural Machine Translation
}

\author{
  Mingbo Ma \thanks{\quad Equal contribution}  \hspace{0.1mm} \thanks{\quad Current address: Baidu Research, Sunnyvale, CA.} $^{\hspace{0.5mm} \dagger}$
  \qquad
  Renjie Zheng $^{* \hspace{0.5mm} \dagger}$
  \qquad
  Liang Huang $^{\dagger,\ddagger}$
\\ 
$^{\dagger}$School of EECS, Oregon State University, Corvallis, OR \\
$^{\ddagger}$Baidu Research, Sunnyvale, CA\\
  {\tt \{cosmmb, zrenj11, liang.huang.sh\}@gmail.com} \\
  }

\date{}

\begin{document}
\maketitle
\begin{abstract}


Beam search optimization \cite{wiseman+rush:2016} 
resolves many issues in neural machine translation.
However, this method lacks principled
stopping criteria and 
does not learn how to stop during training,
and the model naturally prefers {\em{longer}} hypotheses 
during the testing time in practice
since they use the raw score instead of the probability-based score.
We propose a novel ranking method which
enables an optimal beam search stopping criteria.
We further introduce a structured prediction loss function
which penalizes suboptimal finished candidates produced by
beam search during training.
Experiments of neural machine translation on both synthetic
data and real languages (German$\rightarrow$English and Chinese$\rightarrow$English)
demonstrate our proposed methods
lead to better length and BLEU score.

\end{abstract}

\input{intro}

\input{model}

\input{learn}

\input{exps}

\vspace{-3mm}
\section{Future Works and Conclusions}
\vspace{-3mm}

Our proposed methods are general techniques which also can be 
applied to the Transformer \cite{vaswani+:2017}.
As part of our future works, we plan to 
adapt our techniques to the Transformer 
to further evaluate our model's performance.

There are some scenarios that
decoding time beam search is not applicable, such as 
the simultaneous translation system proposed 
by \namecite{Ma2018STACLST} which 
does not allow for adjusting the committed words,
the training time beam search still will be helpful to the 
greedy decoding performance.
We plan to further investigate the performance of 
testing time greedy decoding with beam search optimization during training.

We propose two modifications to BSO to provide better scoring function and 
under-translation penalties, which improves the accuracy in De-En and Zh-En
by 0.8 and 3.7 in BLEU respectively.


\section*{Acknowledgments}

This work was supported in part by DARPA grant N66001-17-2-4030 (M.~M.)
, and NSF grants IIS-1656051 and IIS-1817231 (R.~Z.).


\bibliography{main}
\bibliographystyle{acl_natbib}

\appendix

\end{document}

%% file: defs.tex
\newcommand{\notes}[1]{}

\theoremstyle{definition}

\theoremstyle{plain}

\newcommand{\ith}[1]{\ensuremath{i^{{th}}}}

\newcount\permx
\newcount\permy
\def\permdot#1#2{
\permx=#1 \advance\permx by-1
\permy=#2 \advance\permy by-1
\psframe[fillcolor=black, fillstyle=solid]
(\permx,\permy)(#1, #2)
}

\newcommand{\boxnum}[1]{{\setlength{\fboxsep}{1pt}\raisebox{1pt}{\hspace{1pt}\fbox{\tiny #1}\hspace{1pt}}}}
\newcommand{\ind}[1]{\ensuremath{_{\kern-0.5pt\boxnum{#1}}}}

\newcommand{\vecx}{\ensuremath{\mathbf{x}}\xspace}
\newcommand{\vecy}{\ensuremath{\mathbf{y}}\xspace}

\def\namecite{\newcite}

\newcommand{\smallnt}[1]{\ensuremath{_{\mbox{\tiny PP}}}\xspace}

\newcommand{\pseudocode}{Algorithm}
\floatname{algorithm}{\pseudocode}

\iffalse

\else

\fi

\newcommand{\eos}{\mbox{\scriptsize \texttt{</eos>}}\xspace}

%% file: intro.tex
\section{Introduction}

Sequence-to-sequence (seq2seq) models based on 
RNNs \cite{sutskever+:2014, bahdanau+:2014a}, 
CNNs \cite{Gehring:2017} and 
self-attention \cite{vaswani+:2017}
have achieved 
great successes in Neural Machine Translation (NMT).  
The above family of models encode the source sentence and 
predict the next word 
in an autoregressive fashion 
at each decoding time step. 
The classical  ``cross-entropy'' training objective of seq2seq models is to 
maximize the likelihood of each word in the translation reference 
given the source sentence and all previous words in that reference. 
This word-level loss 
ensures efficient and scalable training of seq2seq models.

However, this word-level training objective 
suffers from a few crucial limitations,
namely the {\em{label bias}} \cite{murray2018correcting}, 
the {\em{exposure bias}}, 
and the {\em{loss-evaluation mismatch}} \cite{lafferty+:2001,Scheduledbengio:2015,Venkatraman:2015ImprovingMP}.
In addition, more importantly, at decoding time,
{\em beam search} is
universally adopted to improve the search quality,
while training is fundamentally local and greedy.
Several researchers have proposed different approaches 
to alleviate above problems, such as reinforcement 
learning-based methods \cite{ranzato+:2016,rennie2017self,zheng+:2018}, 
training with alternative references \cite{shen+:2016,refzheng+:2018}.
Recently, \namecite{wiseman+rush:2016} attempt to address
these issues with a structured training method, Beam Search Optimization (BSO).
While BSO outperforms other proposed methods on German-to-English translation,
it also brings a different set of problems as partially discussed in \cite{ma:2018phdthesis} which we present with details below. 

\begin{figure}[!t]
\centering
\includegraphics[width=7.5cm]{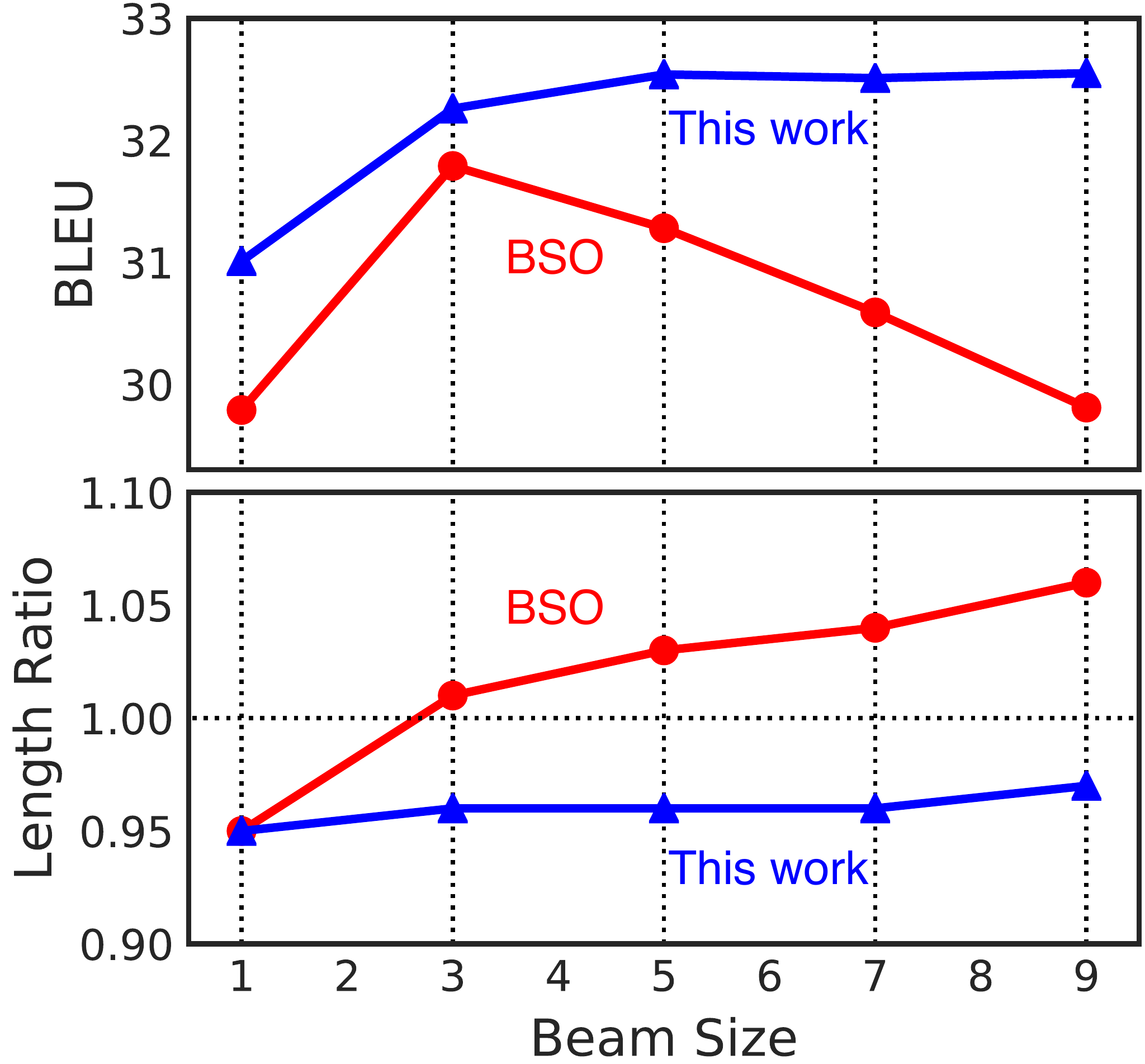}
\vspace{-3mm}
\captionof{figure}{ The BLEU score of BSO decreases after beam size 3 as results of increasing length ratio\footnotemark in German$\rightarrow$English translation. Our model gets higher BLEU with larger beam.}
\label{fig:introbleu}
\vspace{-2mm}
\end{figure}

\footnotetext{There are two types of ``length ratios'' 
in this paper: (a) target to reference ratio ($|\vecy|/|\vecy^*|$),
which is used in BLEU,
and (b)
target to source ratio ($|\vecy|/|\vecx|$). 
By default, the term ``length ratio'' in this paper refers to the former.}

BSO relies on unnormalized raw scores instead of 
locally-normalized probabilities
to get rid of 
the label bias problem.
However, since the raw score can be either positive or negative,
the optimal stopping criteria \cite{huang+:2017} no longer holds, 
e.g., one extra decoding step would increase the entire unfinished 
hypothesis's model score when we have positive word score.
This leads to two consequences: we do not know when to stop the beam search and
it could return overlength translations (Fig.~\ref{fig:introbleu})
or underlength translations (Fig.~\ref{fig:syn}) in practice.
As shown in Fig.~\ref{fig:introbleu}, the BLEU score of BSO drops 
significantly when beam size gets larger as a result 
of overlong translations (as evidenced by length ratios larger than 1). 
Furthermore, BSO performs poorly (shown in Section~\ref{sec:exp}) 
on hard translation pairs, e.g., Chinese$\rightarrow$English (Zh$\rightarrow$En) translation, 
when the target / source ratio is 
more diverse (Table~\ref{tab:dataset}). 


To overcome the above issues, we propose to use the sigmoid function 
instead of the raw score at each time step to rank candidates.
In this way, the model still has probability properties to hold 
optimal stopping criteria without label bias effects.
Moreover, we also encourage the model to generate the hypothesis which is 
more similar to gold reference in length. 
Compared with length reward-based methods \cite{huang+:2017,yilin+:2018},
our model does not need to tune the predicted length and per-word reward.
Experiments on both synthetic and real language translations 
(De$\rightarrow$En and Zh$\rightarrow$En) demonstrate
significant improvements in BLEU score over strong baselines 
and other methods.

%% file: model.tex

\section{Preliminaries: NMT and BSO}
\label{sec:prelim}

Here we briefly review the conventional NMT and 
BSO \cite{wiseman+rush:2016}
to set up the notations. For simplicity, we choose to use RNN-based
model but our methods can be easily applied to other designs of seq2seq 
model as well.

Regardless of the particular design of different seq2seq models,
generally speaking, the decoder always has the following form:
\vspace{-2mm}
\begin{equation}
p(\vecy \mid \vecx) = \textstyle\prod_{t=1}^{|\vecy|}  p(y_t \mid \vecx,\, \vecy_{<t})
\label{eq:gensentscore}
\end{equation}
where $\vecx \in R^{N \times D}$  represents the $D$-dimension hidden 
states from encoder with N words and $\vecy_{<t}$ denotes the gold 
prefix $(y_1,...,y_{(t-1)})$ before t. The conventional NMT model 
is locally trained
to maximize the above probability.

Instead of maximizing each gold word's probability, 
BSO tries to promote the non-probabilistic scores of gold sequence 
within a certain beam size $b$. 
BSO removes the softmax layer and directly uses the raw score after 
hidden-to-vocabulary layer, and the non-probabilistic scoring function 
$f_{\vecx}(y_t \mid \vecy_{<t})$ represents the score of word $y_t$ 
given gold prefix $\vecy_{<t}$ and $\vecx$.
Similarly, $f_{\vecx}(\hat{y}^b_t \mid \hat{\vecy}^b_{<t})$ is 
the $b^{th}$ sequence
with beam size $b$ at time step $t$.
Then, we have the following loss function to penalize the $b^{th}$
candidate and promote gold sequence:
\begin{equation}
\resizebox{.8\hsize}{!}{\hspace{-2mm}
$\mathbb{L} = \displaystyle{\sum_{t=1}^{|\vecy|}} \Delta (\hat{\vecy}^b_{\leq t}) (1+f_{\vecx}(\hat{y}^b_t \mid \hat{\vecy}^b_{<t})-f_{\vecx}(y_t \mid \vecy_{<t}))^+$
\label{eq:bsoloss}}
\end{equation}
where $\Delta (\hat{\vecy}^b_{\leq t})$ is defined 
as $(1-\text{BLEU}(\hat{\vecy}^b_{\leq t},\vecy_{\leq t}))$ 
which scales the loss according to BLEU score between gold and 
$b^{th}$ hypothesis in the beam. 
The notation $(\cdot)^+$ represents a max function
between any value and $0$, i.e., $z^+=max(0,z)$.

When Eq.~\ref{eq:gensentscore} equals to $0$ at time step $t$, 
then the gold sequence's score is higher than the 
last hypothesis in the beam 
by 1, and a positive number otherwise.
Finally, at the end of beam search ($t=|\vecy|$), BSO requires 
the score of $\vecy$ exceed the score of the highest incorrect
hypothesis by $1$.

Note that the above non-probabilistic score function 
$f_{\vecx}(\cdot)$ is not bounded as probabilistic score in conventional NMT.
In practice, when we have positive word score, then the 
unfinished candidates always get higher model scores with 
one extra decoding step and the optimal stopping 
criteria
\footnote{Beam search stops when 
the score of the top unfinished hypothesis is lower than any finished hypothesis, or the \eos is the highest score candidate in the beam.} 
\cite{huang+:2017} is no longer hold.
BSO implements a similar ``shrinking beam'' strategy which duplicates 
top unfinished candidate to replace finished hypotheses and terminates
the beam search when there are only \eos in the beam.
Non-probabilistic score function works well in 
parsing and Statical MT where we know when to stop beam search.
However, in the NMT scenario, without optimal stopping criteria,
we don't know when to stop beam search.


%% file: learn.tex
\section{Learning to Stop}
\label{sec:model}
We propose two major improvements to BSO. 
\subsection{Sigmoid Scoring Function}

As mentioned in Section~\ref{sec:prelim}, BSO relies on raw score function
to eliminate label bias effects. 
However, without using locally-normalized 
score does not mean that we should stop using the probabilistic value function.
Similar with multi-label classification in \cite{Ma2017GroupSC}, 
instead of using locally normalized softmax-based score and 
non-probabilistic raw scores,
we propose to use another form of probabilistic scoring 
function, sigmoid function, which is defined as follows:
\vspace{-2mm}
\begin{equation}
g_{\vecx}(y_t \mid \vecy_{<t}) = (1+e^{ w \cdot f_{\vecx}(y_t \mid \vecy_{<t})})^{-1}
\label{eq:sigmoid}
\vspace{-2mm}
\end{equation}
where $w$ is a trainable scalar parameter which shifts the 
return value of $f_{\vecx}(y_t \mid \vecy_{<t})$ into a non-saturated 
value region of sigmoid function.
Eq.~\ref{eq:sigmoid} measures the probability of each word independently which
is different from locally-normalized softmax function.
Similar to the scenario in multi-label classification,
$g_{\vecx}(y_t \mid \vecy_{<t})$ only promotes the words which are 
preferred by 
gold reference and does not degrade other words. 
Eq.~\ref{eq:sigmoid} enables the model to keep the probability nature
of scoring function without introducing label bias effects.
After the model regain probability-based scoring function,
the optimal stopping criteria can be used in testing time decoding.

\begin{figure}[!t]
\centering
\includegraphics[width=8cm]{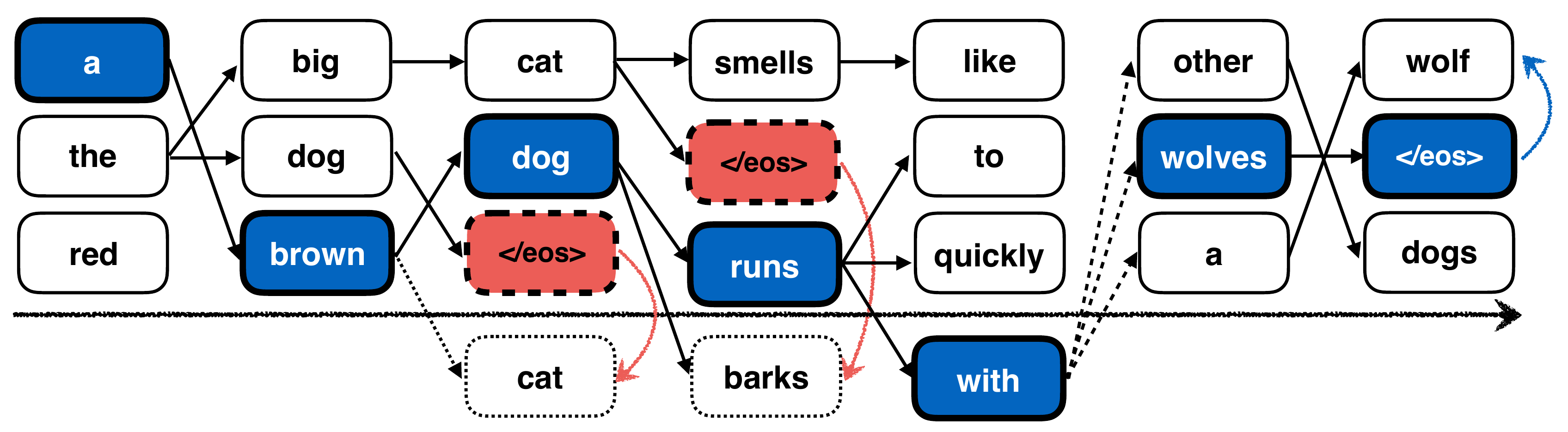}
\captionof{figure}{Training illustration with beam size $b=3$ and gold reference ``a brown dog runs with the wolves''. The gold reference is highlighted in blue solid boxes.  We penalize the under-length translation (short) hypotheses by expelling out the early \eos out of beam (red dashed boxes). The beam search restarts with gold when gold falls off the beam (at step 5).}
\label{fig:model}
\vspace{-3mm}
\end{figure}

\subsection{Early Stopping Penalties}

Similar to Eq.~\ref{eq:gensentscore}, testing time decoder
multiplies the new word's 
probabilistic score with prefix's score when there is a new word appends to 
an unfinished hypothesis.
Though the new word's probabilistic score is upper bounded by 1, 
in practice, the score usually far less than one. 
As described in \cite{huang+:2017,yilin+:2018}, decoder always
prefers short sentence when we use the probabilistic score function.

To overcome the above so-called ``{\em{beam search curse}}'', 
we propose to penalize early-stopped hypothesis within the beam 
during training.
The procedure during training is illustrated in 
Fig.~\ref{fig:model}.

Different from BSO, to penalize the underlength finished translation hypotheses, 
we include additional violations when there is an \eos within 
the beam before the gold reference finishes and we force 
the score of that \eos lower than the $b+1$ candidate by a margin.
This underlength translation violation is formally defined as follows:
\vspace{-1mm}
\begin{equation}
\resizebox{.85\hsize}{!}{
$\begin{split}
&\mathbb{L}^{\text{s}} \!= \!\sum_{t=1}^{|\vecy|} \sum_{j=1}^{b} \mathds{1}(\hat{y}_t^j=\eos) \cdot Q (\hat{y}_t^j,\hat{y}_t^{b+1})\;\;, \\
&Q(\hat{y}_t^j,\hat{y}_t^{b+1}) = ( 
g_{\vecx}(\hat{y}^{j}_t \mid \hat{\vecy}^{j}_{<t}) 
 -g_{\vecx}(\hat{y}^{b+1}_t \mid \hat{\vecy}^{b+1}_{<t}))^+
\end{split}$
\label{eq:eosscore}}
\vspace{-1mm}
\end{equation}
where notation $\mathds{1}$ is identification function which only equals to
1 when $i^{th}$ candidate in beam $\hat{y}_t^j$ is \eos, 
e.g. in Fig.~\ref{fig:model}. 
We only have non-zero loss when the model score of 
underlength translation candidates are greater than the $b+1$ candidate by a margin.
In this way, we penalize all the short hypotheses during training time.
Note that during both training and testing time, 
the decoder stops beam search 
when it satisfies the optimal stopping criteria \cite{huang+:2017}.
Therefore, we do not need to penalize the overlength translations
since we have already promoted the gold reference to the top of the beam at time step $|y|$ during training.

%% file: exps.tex

\vspace{-1mm}
\section{Experiments}
\label{sec:exp}
\vspace{-1mm}

\begin{table}[t]
\centering\setlength{\tabcolsep}{3pt}
\resizebox{0.4\textwidth}{!}{
\begin{tabular}{|l|c|c|c|c|c|c|} \hline
\raisebox{-0.1cm}{Data} & \raisebox{-0.1cm}{Split} & \raisebox{-0.1cm}{$\overline{|x|}$} & \raisebox{-0.1cm}{$\sigma(|x|)$} & \raisebox{-0.2cm}{$\overline{(\frac{|y|}{|x|})}$} & \raisebox{-0.1cm}{$\sigma(\frac{|y|}{|x|})$} & \raisebox{-0.1cm}{\# sents}  \\ \hline
\multirow{3}{*}{Synthetic}& Train &9.47 & 5.45          & 3.0 & 0.52 & 5K\\ \cline{2-7}
        & Valid & 9.54     & 5.42                   & 3.0     & 0.53 & 1K   \\ \cline{2-7}
        & Test  & 9.51     & 5.49                  & 3.0      & 0.52 & 1K  \\ \hline
\multirow{3}{*}{De$\rightarrow$En}& Train &17.53 & 9.93    & 1.07     & 0.16 & 153K   \\ \cline{2-7}
        & Valid & 17.55     & 9.97           & 1.07         & 0.16 & 7K   \\ \cline{2-7}
        & Test$^*$  & 18.89     & 12.82     & 1.06         & 0.16 & 6.5K  \\ \hline
\multirow{3}{*}{Zh$\rightarrow$En} & Train & 23.21 & 13.44 & 1.30 & 0.33 & 1M   \\ \cline{2-7}
        & Valid & 29.53     & 16.62    & 1.34         & 0.22 & 0.6K \\ \cline{2-7}
        & Test  & 26.53     & 15.99    & 1.4          & 0.24 & 0.7K  \\ \hline
\end{tabular}
}
\caption{Dataset statistics of source sentence length and the ratio
between target and source sentences. $\sigma$ is standard deviation. $^*$shows statistics of cleaned test set.}
\label{tab:dataset}
\vspace{-3mm}
\end{table}
We showcase the performance comparisons over three different datasets.
We implement seq2seq model, BSO and our proposed model
based on PyTorch-based OpenNMT \cite{klein+:2017}.
We use a two-layer bidirectional LSTM as the encoder and a two
layer LSTM as the decoder.
We train Seq2seq model for 20 epochs to minimize perplexity on the
training dataset, with a batch size of 64, word embedding size of 512,
the learning rate of 0.1, learning rate decay of 0.5 and dropout 
rate of 0.2.
Following \namecite{wiseman+rush:2016}, we then train BSO and our model
based on the previous Seq2seq model
with the learning rate of 0.01 and learning rate decay of 0.75, batch size of 40.
Note that our pretrained model is softmax-based, 
and we only replace the softmax layer with the sigmoid layer for
later training for simplicity.
The performance will have another boost when
our pretrained model is sigmoid-based. 
We use Adagrad \cite{duchi+:2011} as the optimizer.

In Zh$\rightarrow$En task, we employ BPE \cite{sennrich+:2015} which
reduces the source and target language vocabulary sizes to 18k and 10k.
Following BSO, we set the decoding beam size smaller than the training beam size 
by 1.

 \vspace{-1mm}
\subsection{Synthetic Task}
 \vspace{-1mm}

\begin{figure}[!t]
\centering
\begin{minipage}[t]{1.0 \linewidth}
\begin{center}
\includegraphics[width=6.5cm]{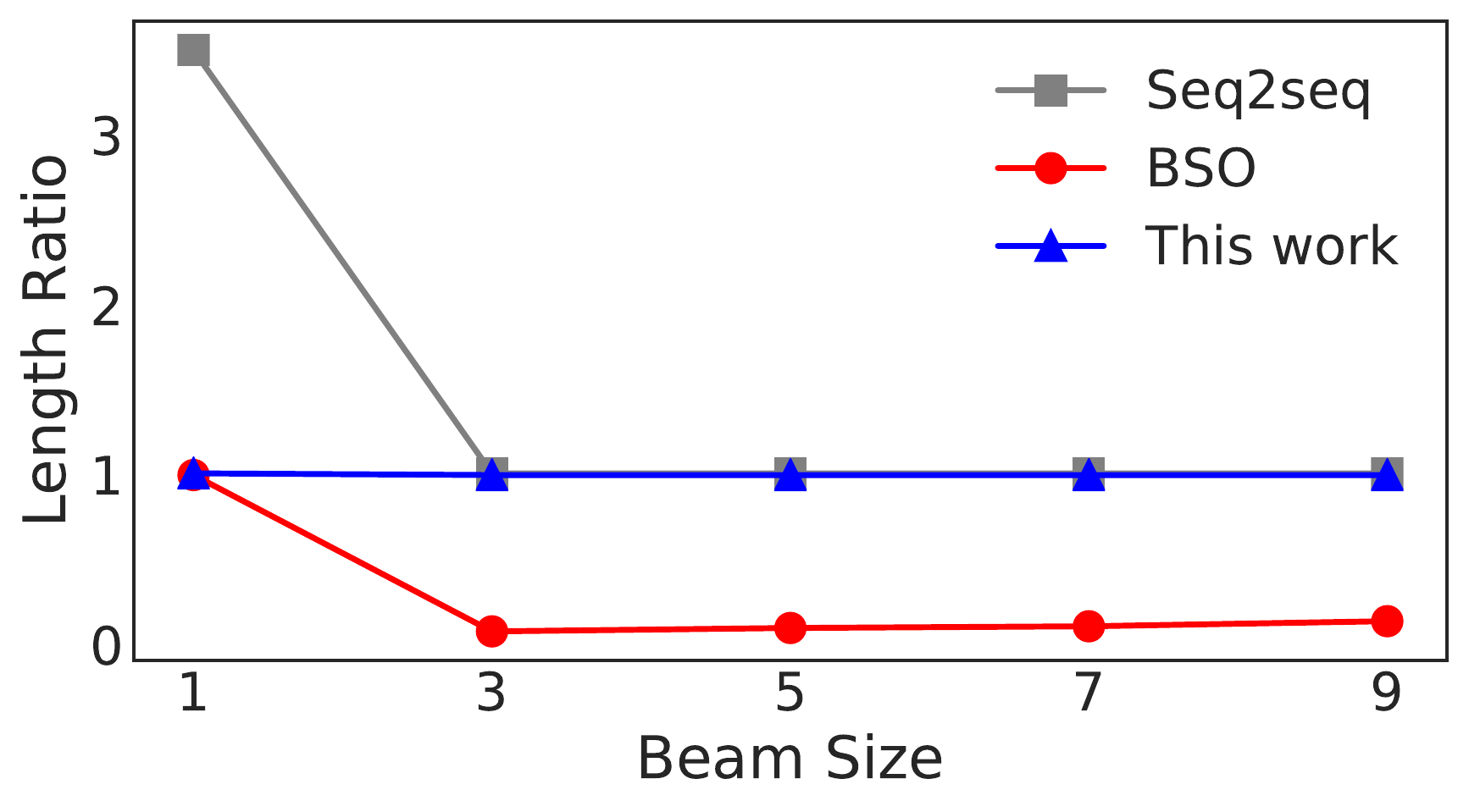}
\end{center}
\end{minipage}
\vspace{-8mm}
\caption{Length ratio on synthetic test dataset.}
\label{fig:syn}
\vspace{-3mm}
\end{figure}

Table \ref{tab:dataset} shows the statistics of 
source sentence length and the ratio between target and source sentences.
The synthetic dataset is a simple translation task which
generates target sentences from this grammar:
$\{a \to x, b \to x\ x, c \to x\ x\ x, d \to x\ x\ x\ x, 
e \to x\ x\ x\ x\ x\}$.
For example:
\begin{enumerate}
\vspace{-1mm}
\item source sentence $[b\ c\ a]$ will generate
the target sentence $[x\ x\ x\ x\ x\ x]$ (2 $x$ from $b$, 3 $x$ from $c$ and
1 $x$ from $a$).
\vspace{-2mm}
\item source sentence $[a, b, c, d, e]$ will be
translated into $[x\ x\ x\ x\ x\ x\ x\ x\ x\ x\ x\ x\ x\ x\ x]$ in target side (1 $x$ from $a$, 2 $x$ from $b$, 3 $x$ from $c$, 4 $x$ from $d$ and 5 $x$ from $e$).
\vspace{-1mm}
\end{enumerate}

This dataset is designed to evaluate the length prediction
ability of different models.
Fig.~\ref{fig:syn} shows the length ratio of different models on the test set.
Only our model can predict target sentence length correctly 
with all beam sizes which shows a better ability to learn target length.

 \vspace{-1mm}
\subsection{De$\rightarrow$En Translation}
 \vspace{-1mm}

\begin{table}[t]
\centering\setlength{\tabcolsep}{3pt}
\resizebox{0.45\textwidth}{!}{
\begin{tabular}{|c|c|c|c|c|c|c|c|} \hline
\small{Decode} & \multicolumn{2}{c|}{Seq2seq$^{\dagger}$ }  & Train &  \multicolumn{2}{c|}{BSO$^{\dagger}$ } & \multicolumn{2}{c|}{This work}    \\ \cline{2-3} \cline{5-8}
\small{Beam}         & BLEU  & Len.& Beam &  BLEU & Len. & BLEU & Len.\\ \hline
1         & 30.65 & 1.00           & 2    & 29.79 & 0.95 & 31.01 & 0.95           \\ \hline
3         & 31.38 & 0.97           & 4    & 31.79 & 1.01 & 32.26 & 0.96           \\ \hline
5         & 31.38 & 0.97           & 6    & 31.28 & 1.03 & 32.54 & 0.96           \\ \hline
7         & 31.42 & 0.96           & 8    & 30.59 & 1.04 & 32.51 & 0.96           \\ \hline
9         & 31.44 & 0.96           & 10   & 29.81 & 1.06 & 32.55 & 0.97           \\ \hline
\end{tabular}
}
\caption{BLEU and length ratio on the De$\rightarrow$En validation set. $^{\dagger}$indicates our own implementation. }
\label{tab:de_dev_result}
\end{table}

\begin{table}[t]
\centering
\resizebox{0.45\textwidth}{!}{
\begin{tabular}{|l|c|c|} \hline
Model                 & BLEU  & Len. \\ \hline
This work (full model)                     & 32.54 & 0.96  \\ \hline
This work  w/ softmax              & 32.29 & 0.98  \\ \hline
This work  w/o scale augment       & 31.97 & 0.95  \\ \hline
This work  w/o early stopping loss & 31.19 & 0.93  \\ \hline
\end{tabular}
}
\caption{Ablation study on the De$\rightarrow$En validation set with training beam size $b=6$.}
\label{tab:ablation}
\vspace{-7mm}
\end{table}

\begin{table}[t]
\centering
\resizebox{0.45\textwidth}{!}{
\begin{tabular}{|l|c|c|c|c|} \hline
\multirow{2}{*}{Model}&\multicolumn{2}{c|}{Original Test Set} &  \multicolumn{2}{c|}{Cleaned Test Set} \\ \cline{2-5}
               & BLEU  & Len. & BLEU  & Len. \\ \hline
BSO$^{\ddagger}$  & 26.35 & -         & \multicolumn{2}{c|}{-} \\ \hline
DAD$^{\ddagger}$  & 22.40 & -         & \multicolumn{2}{c|}{-} \\ \hline
MIXER$^{\ddagger}$ & 21.83 & -         & \multicolumn{2}{c|}{-} \\ \hline
Seq2seq$^{\dagger}$      & 29.54 & 0.97      & 30.08 & 0.97     \\ \hline
BSO$^{\dagger}$          & 29.63 & 1.02      & 30.08 & 1.02     \\ \hline
This work                & 30.29 & 0.98      & 30.85 & 0.98      \\ \hline
\end{tabular}
}
\caption{BLEU and length ratio on the De$\rightarrow$En test set. $^{\dagger}$indicates our own implementation. $^{\ddagger}$results from \cite{wiseman+rush:2016}. }
\label{tab:de_test_result}
\end{table}

\begin{figure}[!t]
\centering
\begin{minipage}[t]{1.0 \linewidth}
\begin{center}
\includegraphics[height=6.0cm]{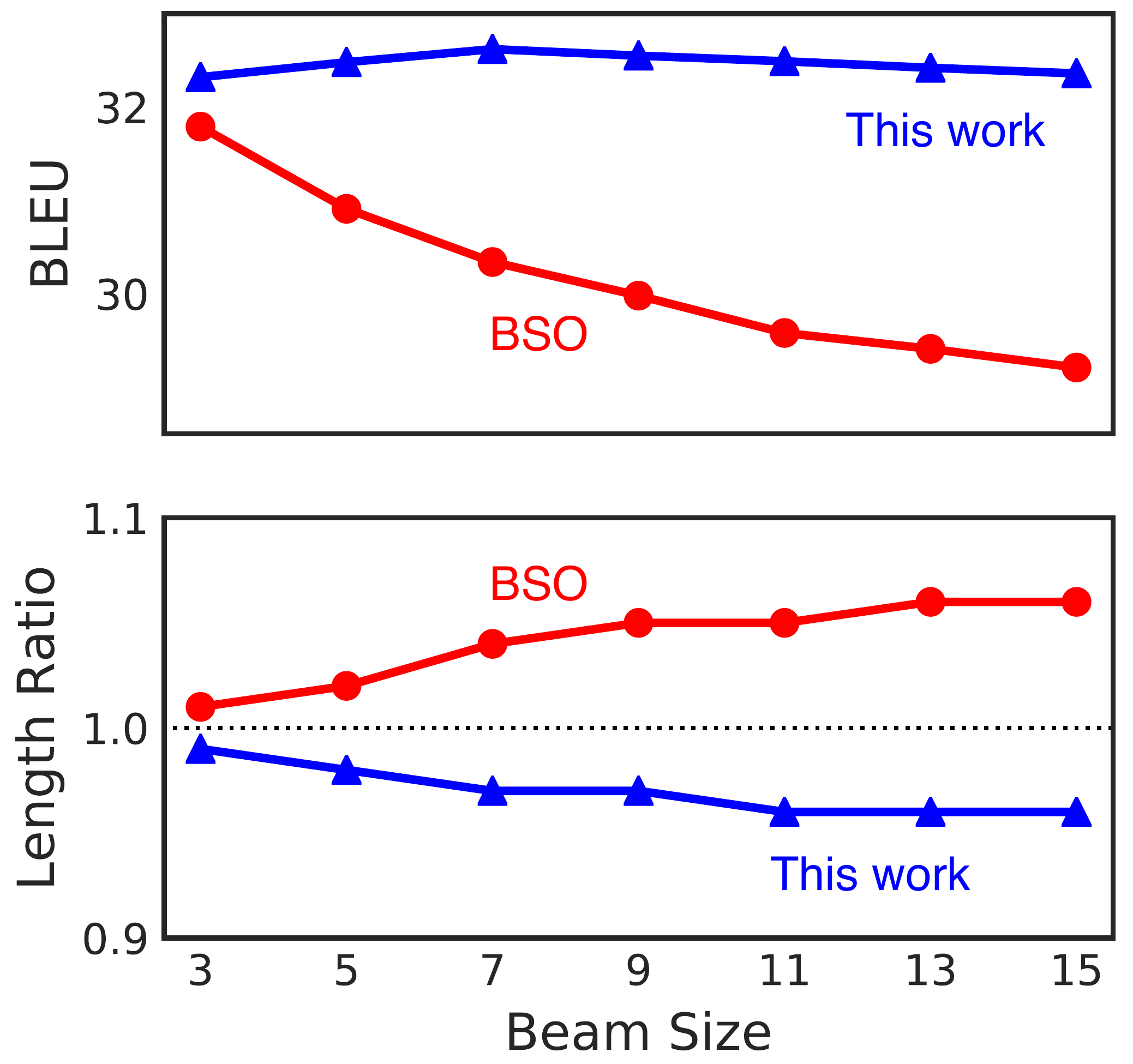}
\end{center}
\end{minipage}
\caption{BLEU and length ratio of models with training beam size $b=6$ and decode with different beam size on De$\rightarrow$En dataset.}
\label{fig:de2en}
\vspace{-10pt}
\end{figure}


The De$\rightarrow$En dataset is previously
used in BSO and MIXER \cite{ranzato+:2016}, which is from IWSLT 2014
machine translation evaluation campaign  
\cite{cettolo+:2014}
\footnote{The test set of De$\rightarrow$En involves some mismatched
source-reference pairs.
We have cleaned this test set and report the statistics
based on the cleaned version.}.

Table \ref{tab:de_dev_result} shows the BLEU score and length ratio of
different models on dev-set. Similar to seq2seq, our proposed model
achieves better BLEU score with larger beam size and outperforms
the best BSO $b=4$ model with 0.76 BLEU.
The ablation study in Table \ref{tab:ablation} shows that
the model produces shorter sentence without scale augment
(term $\Delta (\hat{y}_{\leq t}^{b})$ in Eq.~\ref{eq:bsoloss})
and early stopping loss.
The model also performs worse when replacing softmax to sigmoid
because of the label bias problem.
Fig.~\ref{fig:de2en} shows BLEU score and length ratio of
BSO and our models trained with beam size $b=6$ with
different decoding beam size.
Compared with BSO, whose BLEU score degrades dramatically 
when increasing beam size, our model performs much more stable.
Moreover, BSO achieves much better BLEU score with decoding beam $b=3$ while 
trained with $b=6$ because of a better length ratio,
this is inconsistent with their claim that decoding beam size
should smaller than training beam size by 1.

Table \ref{tab:de_test_result} shows better accuracy of our proposed model
than not only published test results of BSO \cite{wiseman+rush:2016},
DAD \cite{bengio2015scheduled} and MIXER \cite{ranzato+:2016},
but also our implemented seq2seq and BSO model.

 \vspace{-1mm}
\subsection{Zh$\rightarrow$En Translation}
 \vspace{-1mm}

\begin{figure}
\centering
\begin{minipage}{1.0 \linewidth}
\begin{center}
\includegraphics[height=7.2cm]{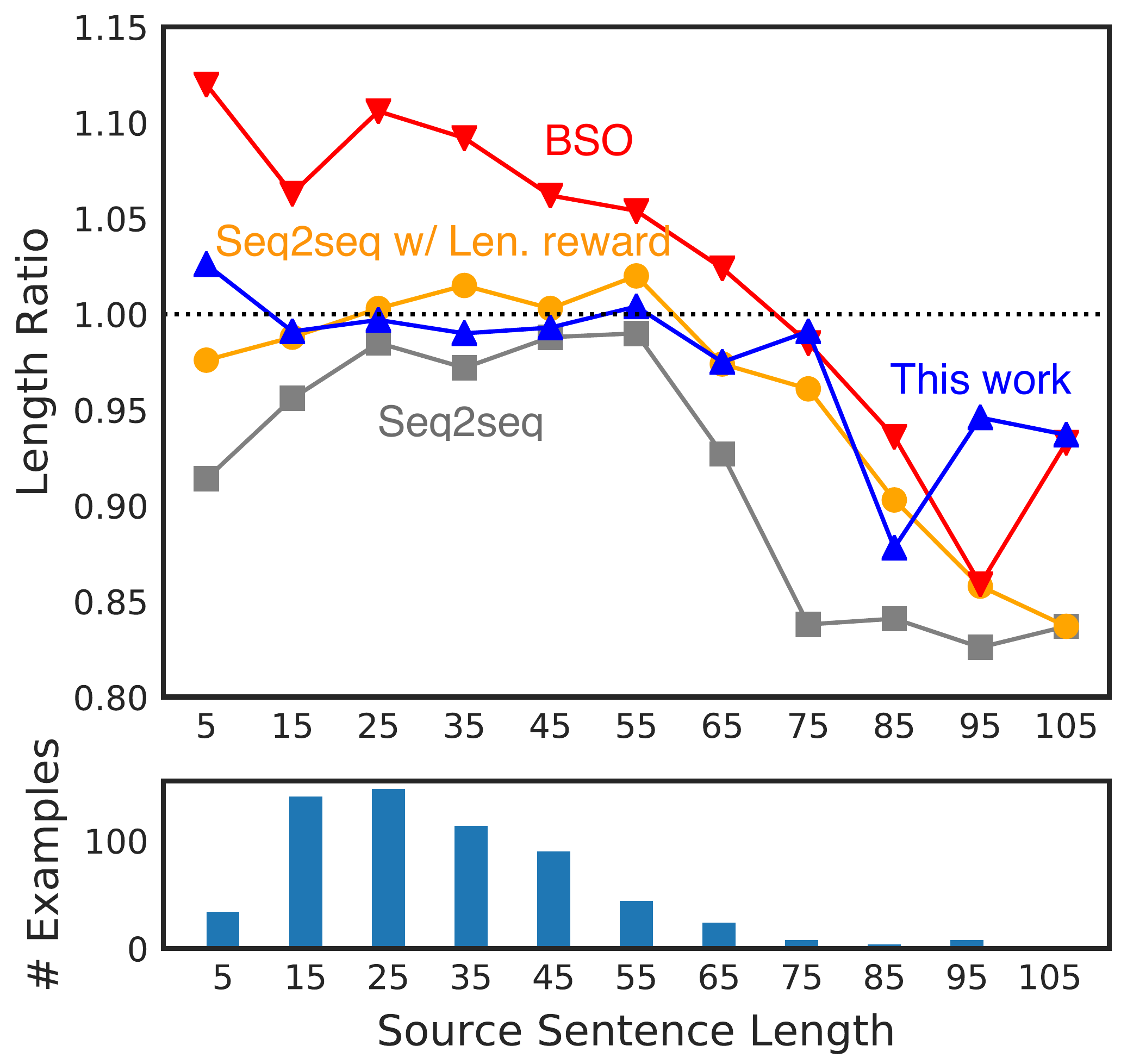}
\end{center}
\end{minipage}
\caption{Length ratio of examples on Zh$\rightarrow$En dev-set with different source sentence length.}
\label{fig:zh2en}
\vspace{-15pt}
\end{figure}

\begin{table}[!htb]
\centering
\resizebox{0.45\textwidth}{!}{
\begin{tabular}{|l|c|c|c|c|} \hline
\multirow{2}{*}{Model} & Train & Decode & \multirow{2}{*}{BLEU}  & \multirow{2}{*}{Len.} \\
         & Beam & Beam  &       &  \\ \hline
Seq2Seq$^{\dagger}$  & -    & 7     & 37.74 & 0.96    \\ 
\: w/ Len. reward$^{\dagger}$ & -    & 7     & 38.28 & 0.99 \\ \hline
BSO$^{\dagger}$& 4    & 3     & 36.91 & 1.03    \\ \hline
BSO$^{\dagger}$& 8    & 7     & 35.57 & 1.07   \\ \hline
This work& 4    & 3     & 38.41 & 1.00   \\ \hline
This work& 8    & 7     & 39.51 & 1.00    \\ \hline
\end{tabular}
}
\caption{BLEU and length ratio of models on Zh$\rightarrow$En validation set.  $^{\dagger}$indicates our own implementation.}
\label{tab:zh_dev_result}
\end{table}

\begin{table}[!htb]
\centering
\resizebox{0.32\textwidth}{!}{
\begin{tabular}{|l|c|c|c|c|} \hline
Model    & BLEU  & Len. \\ \hline
Seq2Seq$^{\dagger}$ & 34.19 & 0.95        \\
\: w/ Len. reward$^{\dagger}$ & 34.60 & 0.99        \\ \hline
BSO$^{\dagger}$   & 31.78 & 1.04        \\ \hline
This work& 35.40 & 0.99       \\ \hline
\end{tabular}
}
\vspace{-2mm}
\caption{BLEU and length ratio of models on Zh$\rightarrow$En test set.  $^{\dagger}$indicates our own implementation.}
\label{tab:zh_test_result}
\vspace{-5mm}
\end{table}

We also perform experiments on NIST Zh$\rightarrow$En translation dataset.
We use the NIST 06 and 08 dataset with 4 references as the validation 
and test set respectively.
Table~\ref{tab:dataset} shows that the characteristic of Zh$\rightarrow$En translation
is very different from De$\rightarrow$En in source length and variance in 
target/source length ratio.

We compare our model
with seq2seq, BSO and seq2seq with length reward \cite{huang+:2017}
which involves hyper-parameter to solve neural model's tendency
for shorter hypotheses
(our proposed method does not require tuning of hyper-parameter).
Fig.~\ref{fig:zh2en} shows that BSO prefers overlength hypotheses
in short source sentences and underlength hypotheses when the source
sentences are long.
This phenomenon degrades the BLEU score in 
dev-set from Table \ref{tab:zh_dev_result}.
Our proposed model comparatively achieves better length ratio 
on almost all source sentence length in dev-set.